\pdfoutput=1

\documentclass[11pt]{article}
\usepackage[]{acl}

\usepackage{times}
\usepackage{latexsym}
\usepackage{graphicx}
\usepackage{subfig}

\usepackage[T1]{fontenc}

\usepackage[utf8]{inputenc}

\usepackage{microtype}

\title{Transformer based ensemble for emotion detection}

\newcommand{\repeatthanks}{\textsuperscript{\thefootnote}}
\newcommand\blfootnote[1]{%
  \begingroup
  \renewcommand\thefootnote{}\footnote{#1}%
  \addtocounter{footnote}{-1}%
  \endgroup
}

\author{Aditya Kane\thanks{~~First authors} \and Shantanu Patankar\repeatthanks \\
  Pune Institute of Computer Technology, Pune\\
  \texttt{\{adityakane1, shantanupatankar2001\}@gmail.com} \\
  \AND Sahil Khose \and Neeraja Kirtane \\
  Manipal Institute of Technology, Manipal \\
  \texttt{\{sahilkhose18, kirtane.neeraja\}@gmail.com} \\}

\begin{document}
\maketitle
\begin{abstract}
    Detecting emotions in languages is important to accomplish a complete interaction between humans and machines. This paper describes our contribution to the WASSA 2022 shared task which handles this crucial task of emotion detection. We have to identify the following emotions: sadness, surprise, neutral, anger, fear, disgust, joy based on a given essay text. We are using an ensemble of ELECTRA and BERT models to tackle this problem achieving an F1 score of $62.76\%$. Our codebase \footnote{\url{https://bit.ly/WASSA_shared_task}} and our WandB project\footnote{\url{https://wandb.ai/acl_wassa_pictxmanipal/acl_wassa}} is publicly available.
\end{abstract}

\blfootnote{\textit{Accepted at the ACL WASSA workshop 2022.}}

\section{Introduction}
Even after engineering a 175B parameter language model like GPT-3 \cite{gpt3} we are far from artificial general intelligence. 
Emotion is a concept that is challenging to describe. However, as human beings, we understand the emotional effect that situations could have on other people. It is interesting to see how we can infuse this knowledge into machines. This work explores whether it is possible for machines to map emotions to situations consciously.
Emotion in text has been studied for a while and has given interesting insights. The dataset that we are using is an extended version of the \citep{ekman_ds} dataset. 
Our team, MPA\_ED, participated in the WASSA 2022 Shared Task on Empathy Detection and Emotion Classification, Track 2: Emotion Classification (EMO), which consists of predicting the emotion at the essay level. This paper has the following contributions:

We propose three new datasets generated using various sampling techniques which overcome the class imbalance. 
We present our ensemble based solution consisting of multiple ELECTRA and BERT \citep{bert} models to solve the emotion classification task. 
We provide a detailed analysis of the performance of the cluster of models and reflect on the shortcomings of the models as well as the dataset generated that affected the performance.


\section{Related Work}

Emotion detection and sentiment analysis has been an extensive research topic since the inception of natural language processing. It has been studied in great detail by faculties of both computer science and neurobiology \citep{cognitive}. \citet{Murthy_2021} presents an extensive review of the modern emotion  classification techniques. The work by \citet{spanemo} remains the current state-of-the-art on emotion classification on the renowned SemEval dataset \citep{semeval}. BERT remains the best performer on the GoEmotions dataset \citep{goemotions}  

\begin{figure}[!tbh]
\centering
\includegraphics[height=4cm]{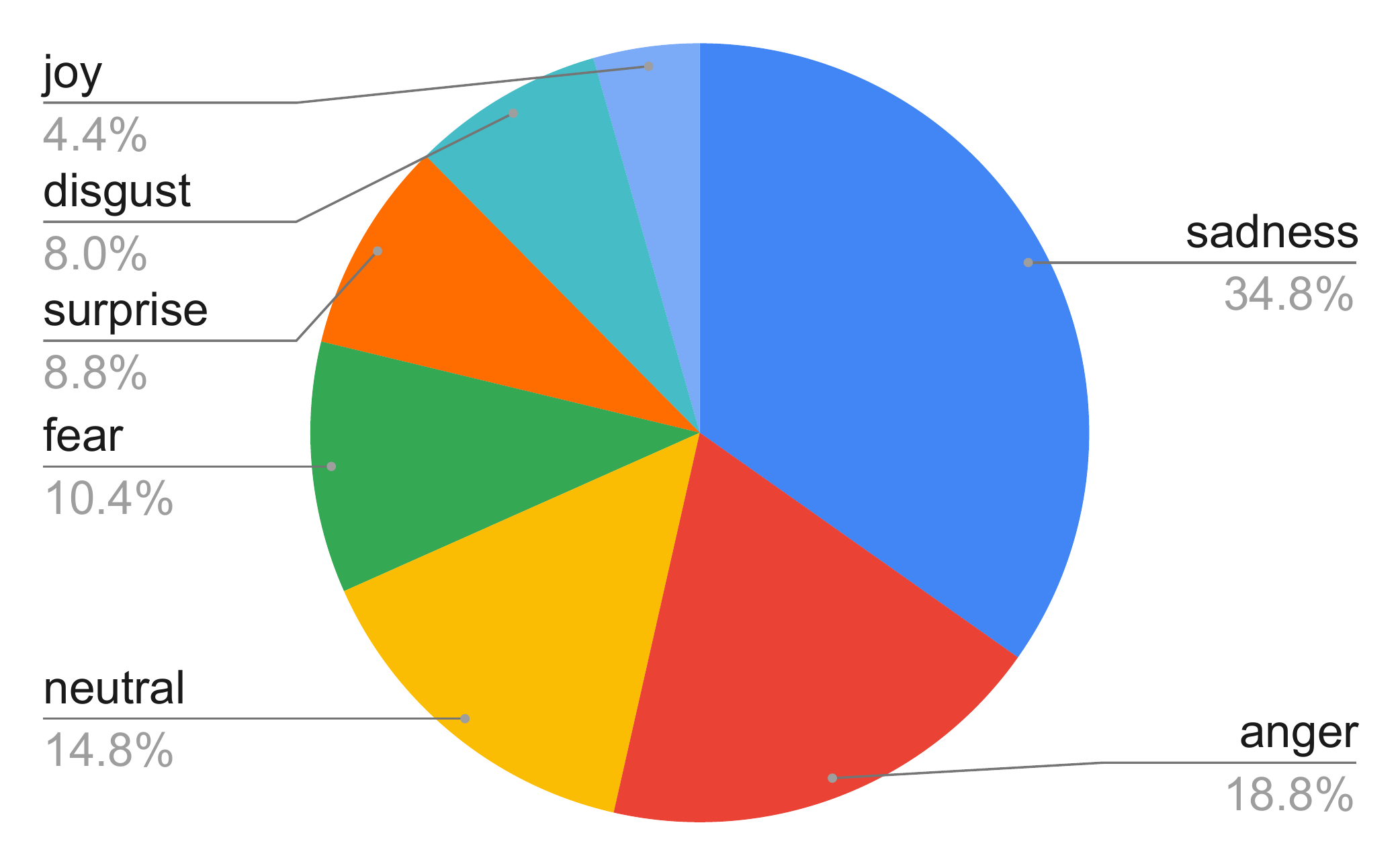}
\caption{Class distribution in emotions}
\label{fig1}
\end{figure}
\section{Data}
\label{Data}
The dataset consists of 1860 data points. Each data point has an essay and its emotion. The emotions are classified into seven types: anger, disgust, 
\begin{figure*}[!tbh]
    \centering
    \includegraphics[height=3cm]{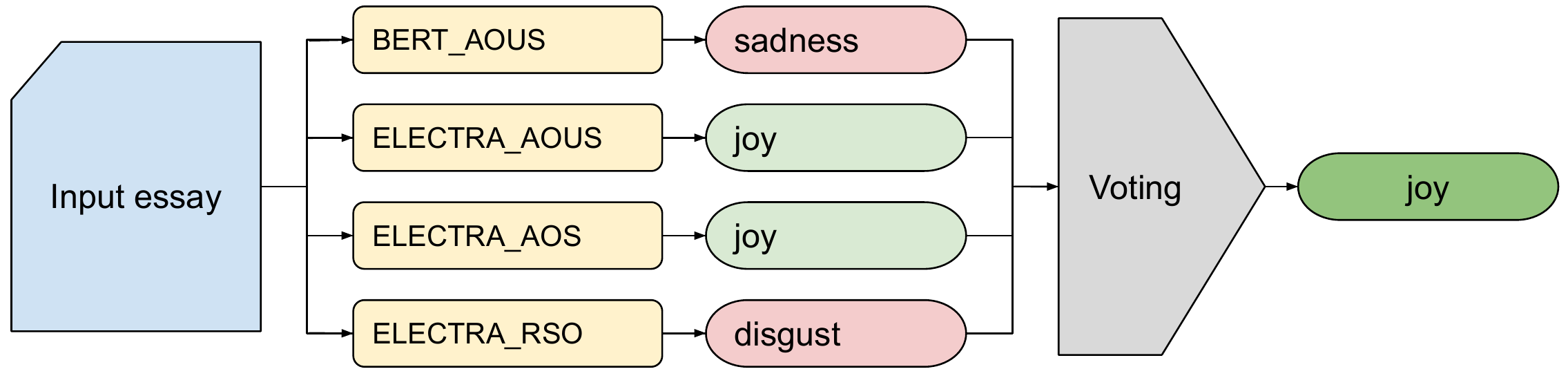}
    \caption{Ensemble pipeline}
    \label{fig:ensemble}
\end{figure*}
fear, joy, neutral, sadness, and surprise. The validation and test split has 270 and 525 data points respectively. The classes for the training data expresses high imbalance, as shown in Fig \ref{fig1} . Here we see that the emotion "sadness" has the maximum number of data points, whereas "joy" has the least number of data points. The distribution is highly skewed and hence data augmentation is required to mitigate that. We performed basic preprocessing like removing punctuation, numbers, multiple spaces, and single line characters. 

\begin{figure}[!t]
    \includegraphics[height=4.5cm]{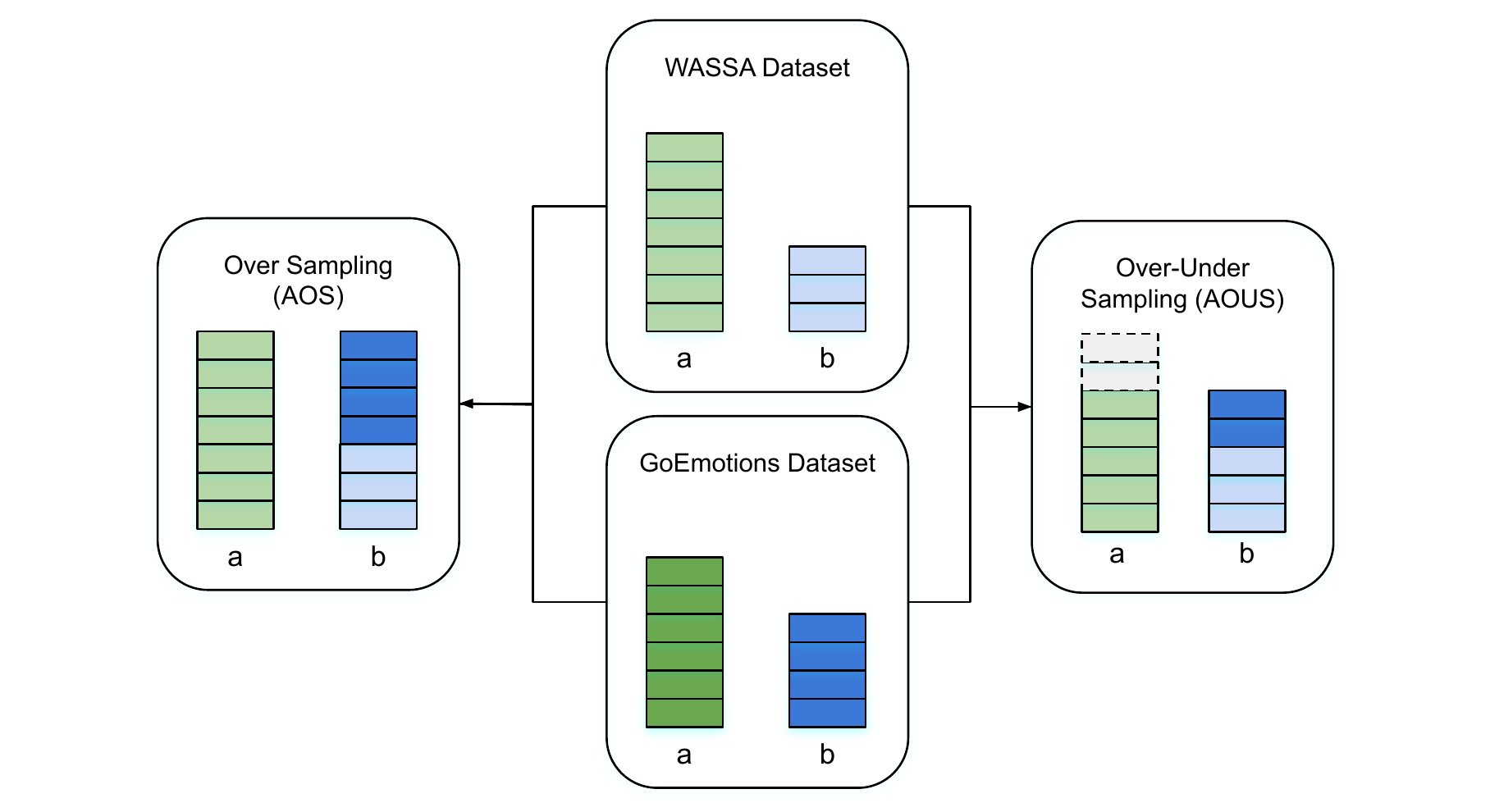}
    \caption{AOS and AOUS}
    \label{fig:aos}
\end{figure}

To overcome the class imbalance, GoEmotions dataset is used, which is a similar dataset with 27 emotions. 
We suggest three data augmentation techniques using the dataset described as follows:
\begin{itemize}


    \item \textbf{Augmented Over-UnderSampling (AOUS)}: If $X$ denotes the number of data points per class, in this method, if the data points in a particular class are greater than $X$, we undersample the data by randomly removing the essays. Otherwise, the data is oversampled by simply adding Reddit comments with maximum lengths from GoEmotions dataset (sorted by lengths) (Fig \ref{fig:aos}). As the average length of comments in GoEmotions dataset is 12 and average length of essays in WASSA dataset is 84, the comments with maximum length are chosen for oversampling. We take $X$ as 400 in our experiments. \vspace{-0.5em}
    
    \item \textbf{Random synthetic oversampling (RSO)}: We observe a significant difference in the average comment length of GoEmotions dataset and the average essay length in the WASSA dataset. To avoid disturbing the length distribution of the WASSA dataset after oversampling, we create synthetic essays by concatenating  multiple random comments with same emotion (Fig \ref{fig:rso}). 
We match the distribution of lengths of the synthetically generated essays from GoEmotions dataset with the distribution of the original dataset using “Systematic Sampling.” We eliminate the deficit in each class by adding synthetically generated essays.
    
    
    \item \textbf{Augmented Oversampling (AOS)}: $X$ denotes the highest number of data points per class. If the number of data points is less than $X$, the data is oversampled by adding comments from GoEmotions dataset with the highest lengths. (Fig \ref{fig:aos})
\vspace{-0.5em}
\end{itemize}
The data distribution post augmentation is balanced with number of samples in AOS, RSO and AOUS datasets equal to 4528, 4828 and 2800 respectively.



\section{System Description}
Bidirectional Encoder Representations from Transformers (BERT) \citep{bert} is a transformer-based \citep{vaswani2017attention} language model developed by Google.  \par
ELECTRA \citep{electra} is a variation of BERT, having a different pre-training approach. It requires less compute time compared to BERT.\par 
We performed ablations with many of the present well-known language models — ALBERT \cite{albert}, XLNET \cite{xlnet}, RoBERTa \cite{roberta} and found BERT and ELECTRA to perform the best.

\section{Ensemble Methods}
We conducted extensive experimentation and observed some models to perform substantially better than others. We shortlisted the models based on the validation F1-score. We decided to ensemble these models for better performance. We shortlisted four models and used majority voting as our ensemble method: BERT with AOUS, ELECTRA with AOS, ELECTRA with RSO, ELECTRA with AOUS.

\begin{table}
\centering
\begin{tabular}{lll}
\hline
\textbf{Model} & \textbf{Dataset} & \textbf{macro F1} \\
\hline
BERT$_{base}$ & AOUS & $59.19\%$ \\
ELECTRA$_{base}$ & AOS & $58.94\%$ \\
ELECTRA$_{base}$ & RSO & $59.06\%$ \\
ELECTRA$_{base}$ & AOUS & $59.67\%$ \\
\hline
\textbf{Ensemble} & val & \textbf{$62.76\%$} \\
 & test & \textbf{$53.41\%$} \\
\hline
\end{tabular}
\caption{Validation metrics}
\label{tab:validation_metrics}
\end{table}





We used the ensemble of the models in Table \ref{tab:validation_metrics}. The confusion matrices of each of these models are shown in Fig \ref{fig:combined_confusion}. The confusion matrix of the resultant ensemble is shown in \ref{fig:ensemble_confusion}. Note that all confusion matrices are normalized by the number of true samples in each class of the evaluation dataset.
 We deduce the following observations:
\begin{enumerate}
    \item When the true label is "disgust," all models confuse the emotions "anger" and "disgust". All models have below average performance on "anger" and "disgust". \vspace{-0.5em}
    \item Models trained on AOUS dataset (c, d in Fig \ref{fig:combined_confusion}) are less prone to confusion in multiple close classes like "disgust", "fear" and "sadness" . \vspace{-0.5em}
    \item The emotions "anger" and "disgust" do not benefit from the ensemble, whereas "fear" suffers a bit.
    However we observe, the emotions "neutral", "sadness" and "surprise" experience significant gains from this process.
    \vspace{-0.5em}
    
\end{enumerate}

\section{Experiments and Results}



Our training setup was fairly straightforward. Language model backbone followed by fully connected layer and Softmax is used. CrossEntropy loss was used. We employed the Adam optimizer with $1e-5$ learning rate and batch size of $8$. We fixed the seed for \verb|numpy| and \verb|torch| to 3407.

Some of the observations made during our extensive experimentation is as follows:

\begin{enumerate}

    \item \textbf{Batch size 8 outperforms larger batch sizes}: We observed improvements across all models 
        \begin{figure}[!h]
    \includegraphics[height=5cm]{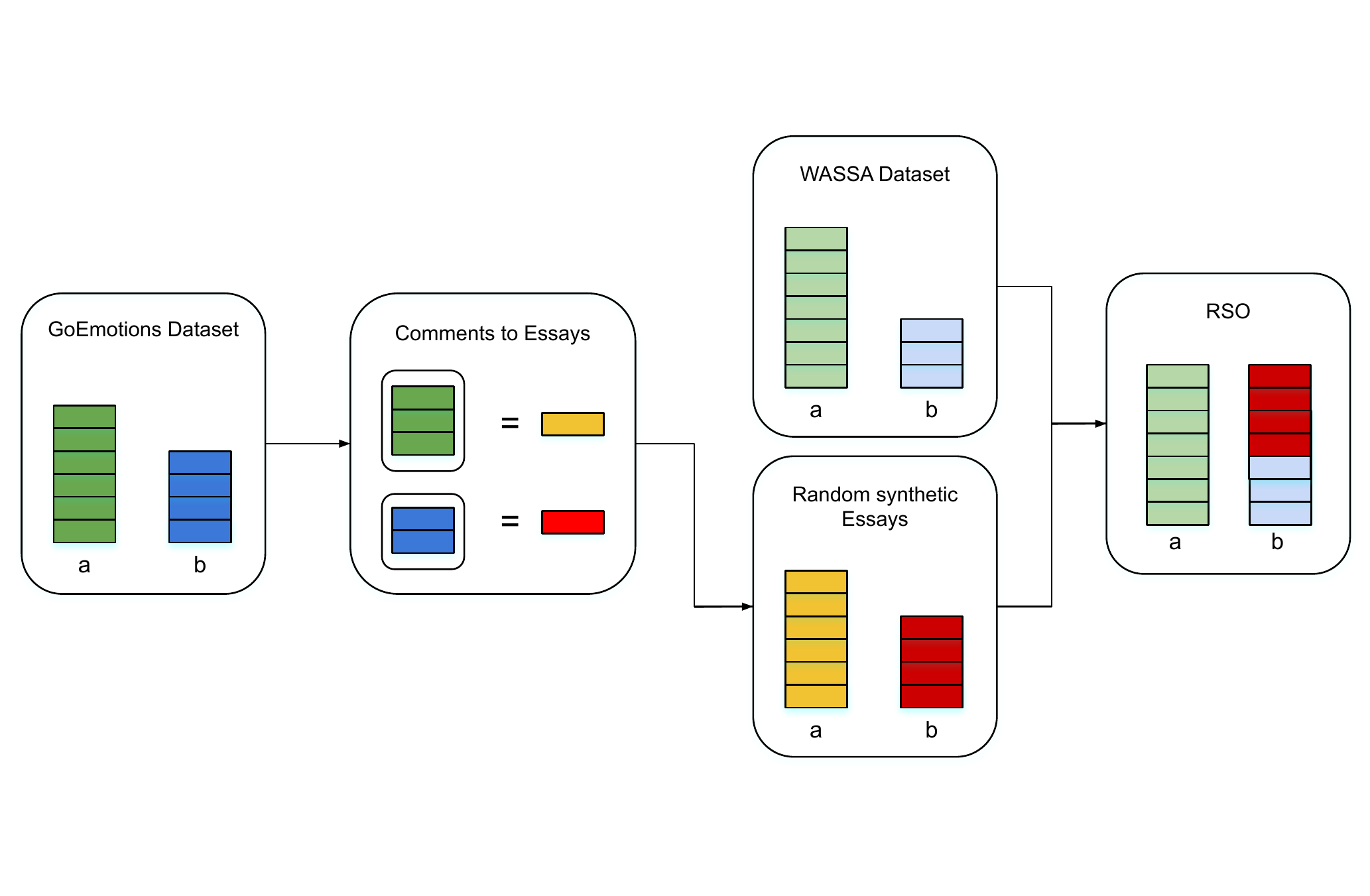}
    \caption{RSO}
    \label{fig:rso}
\end{figure}
    and datasets using a batch size of 8 over 32 or 64. We speculate this is because smaller batch size helps in  generalization as the stochasticity of individual batches increase.
    
    \begin{figure*}[!tbh]
\centering
\begin{tabular}{cc}
  \includegraphics[height=4cm]{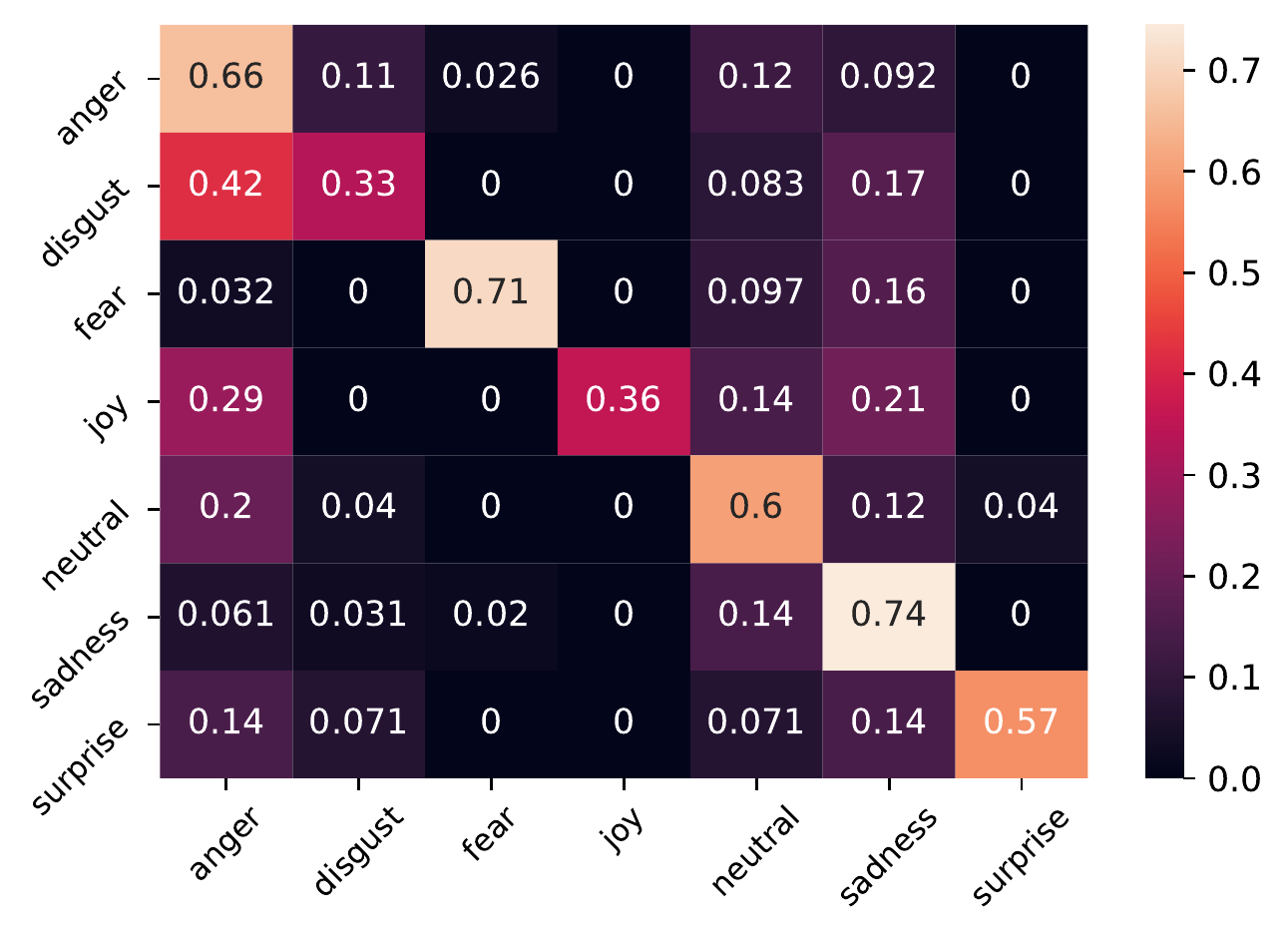} &  
  \includegraphics[height=4cm]{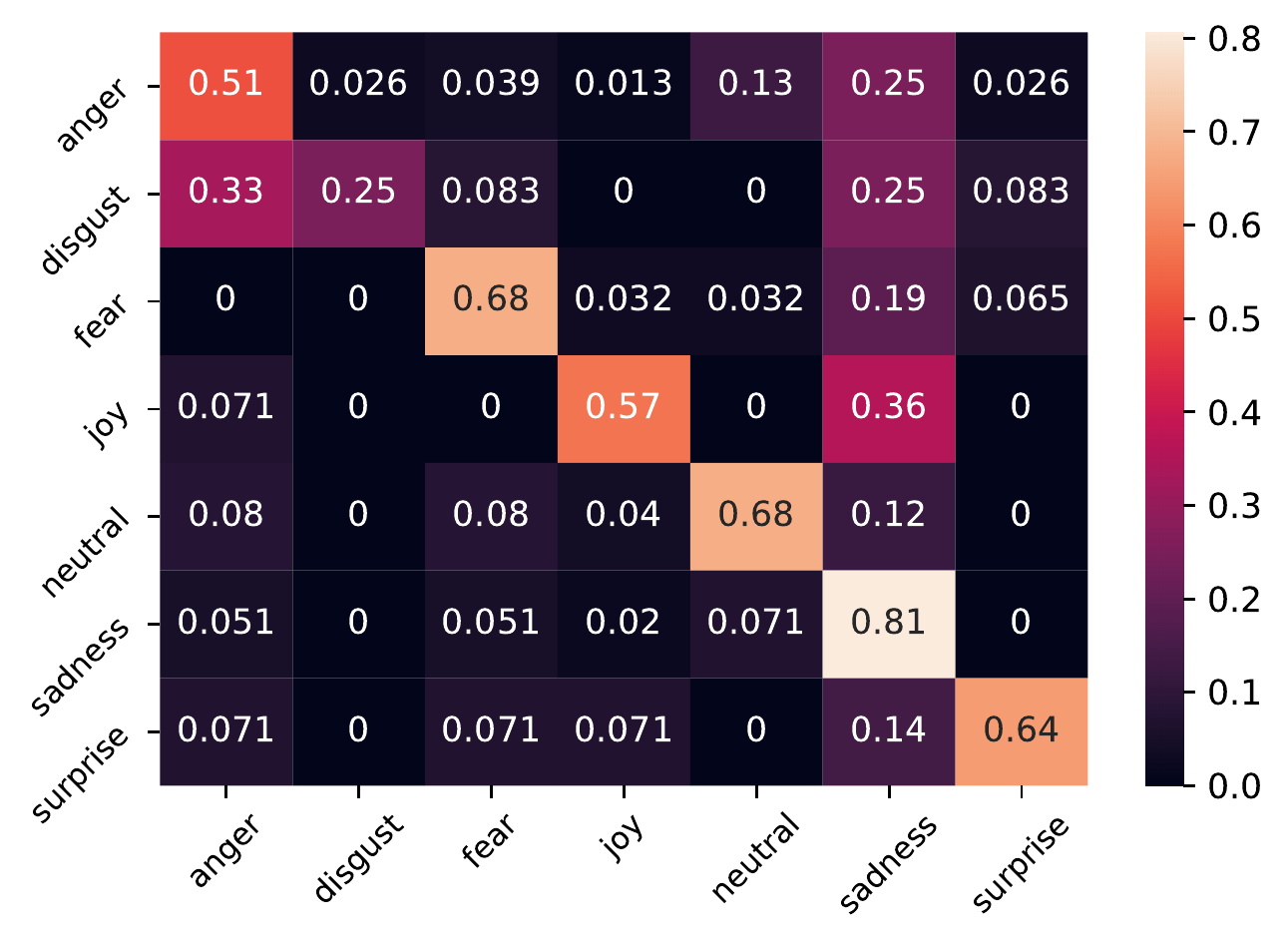} \\
(a) ELECTRA with AOS & (b) ELECTRA with RSO \\
 \includegraphics[height=4cm]{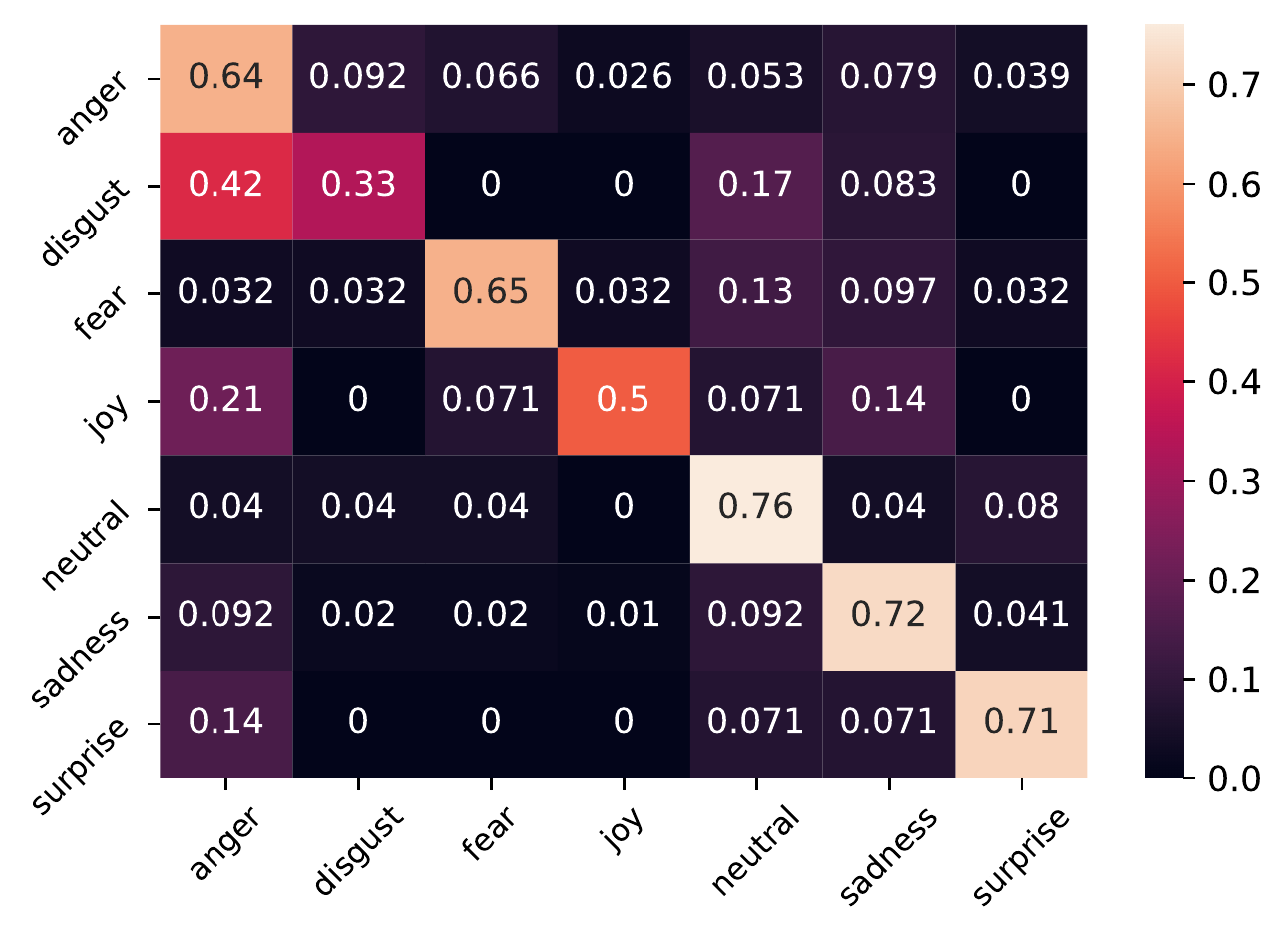} &   \includegraphics[height=4cm]{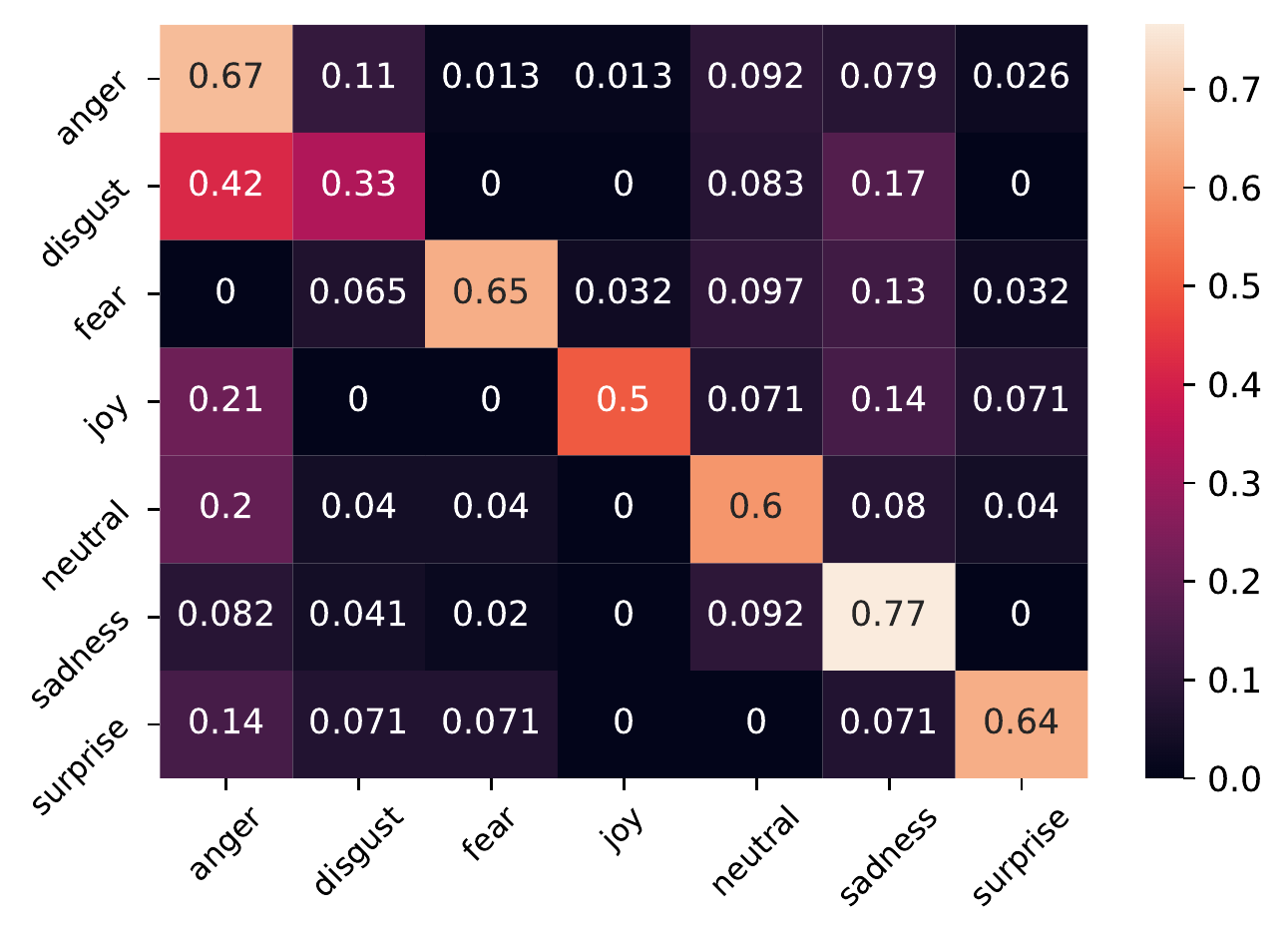}  \\
(c) ELECTRA with AOUS & (d) BERT with AOUS \\[3pt]
\end{tabular}
\caption{Confusion matrices of our models.}
\label{fig:combined_confusion}
\end{figure*}

\begin{figure}[!tbh]
\centering
\includegraphics[height=4cm]{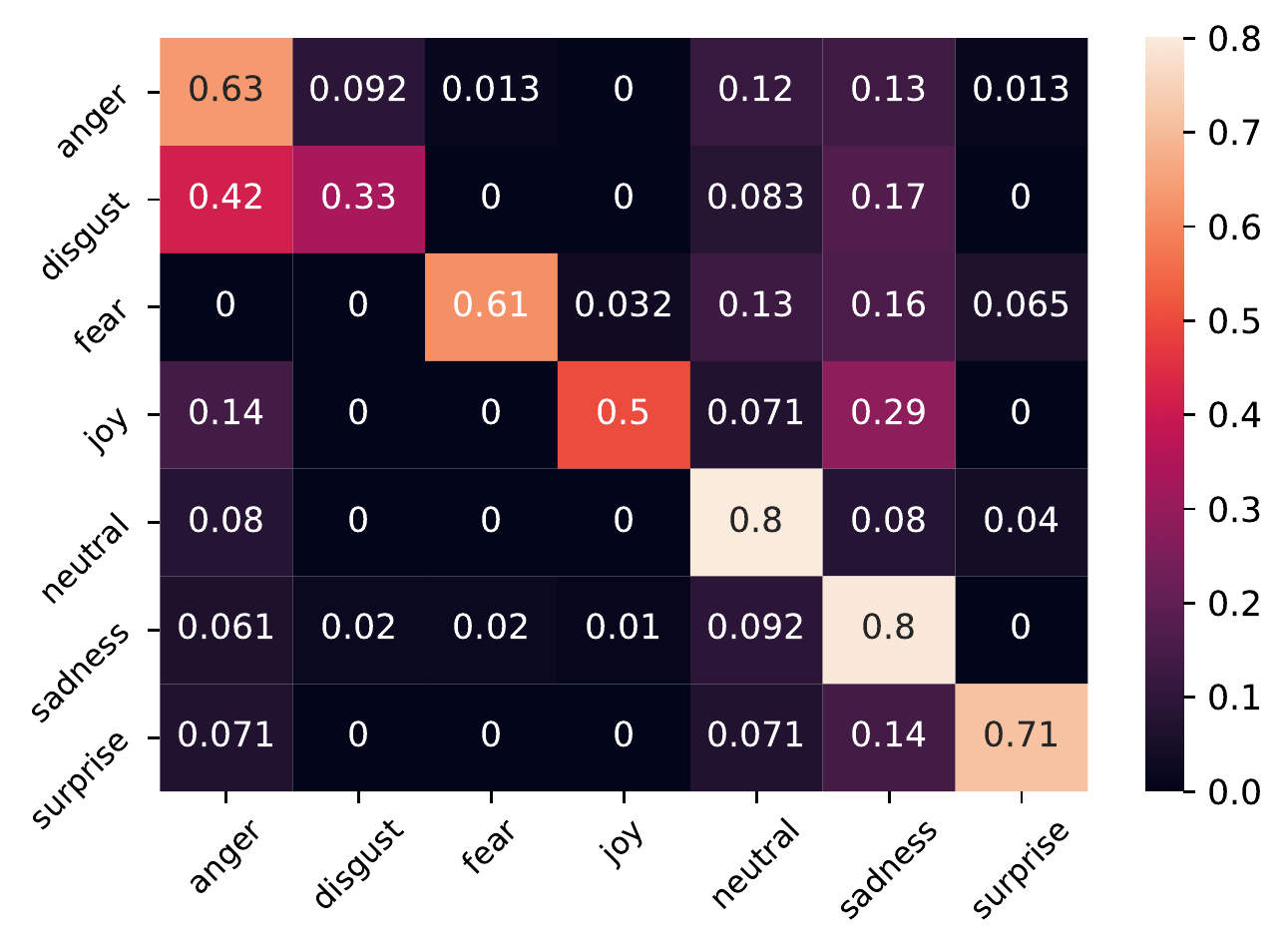}
\caption{Confusion matrix of our final ensemble.}
\label{fig:ensemble_confusion}
\end{figure}

\item \textbf{ELECTRA fine-tuned on the AOUS dataset outperforms other models}: ELECTRA performs better than BERT for all our augmented datasets.  We believe models finetuned on AOUS dataset perform better because AOUS dataset has 400 labels per class, making the dataset balanced while limiting the adulteration induced by the GoEmotions dataset.
    \vspace{-0.5em}
    \item \textbf{Multi-task learning has poor performance}: We experimented with multi-task learning where empathy and distress tasks (Track 1) and emotion classification task (Track 2) were trained together with a shared backbone. We observed that the training was erratic, and the training loss did not converge.
    \item \textbf{Models are sensitive to data imbalance}: When trained on the original dataset with class imbalance, the model is biased towards predicting classes with more training samples. We used data augmentation techniques mentioned in Section \ref{Data} to tackle this issue. After handling the class imbalance with data augmentation, the macro F1 score of the BERT model increased from $32.19\%$ to $59.19\%$. \vspace{-0.5em}

    \item \textbf{Emotion} \textbf{"joy" vs "surprise"}: These are the only two positive emotions in the dataset. We expected all of the models to confuse these emotions as they are semantically similar. However, to our "surprise", we observed the models performed spectacularly on these two emotions. We think this is because "surprise" and "joy" have distinct appearances in the corpus. "surprise" examples have some sort of exclamation or a questioning tone in them. This leaves us with "joy", which happens to be the only positive emotion along with "surprise" in the corpus.  \vspace{-0.5em}

    \item \textbf{Randomly created synthetic essays provide little understanding}: We observed the model trained on RSO augmented data often predicts other emotions as "sadness" (Fig \ref{fig:combined_confusion} (b)). We speculate this is because there was no addition of synthetically generated data for the "sadness" class as it is the largest class. We further hypothesize the synthetic data in RSO, being randomly concatenated, disrupts the context of the entire essay as a whole. However, we still use the model in our final ensemble since it performed well amongst the population. We think this occurs due to multiple factors being simultaneously at play. Further investigation is a promising future direction.
    
\end{enumerate}

The validation confusion matrix of all the four models are displayed in Fig \ref{fig:combined_confusion} and their results in Table \ref{tab:validation_metrics}. We present the following statistics. (True Positive (TP), standard deviation ($\sigma$), mean ($\mu$))

\begin{enumerate}
    \item \textbf{The highest TP $\mu$} is for \textbf{"sadness"} and \textbf{"fear"} emotion with 76 and 67.25 values respectively. Interestingly both of these emotions also have the least TP $\sigma$ with 3.92 and 2.87 values respectively. \vspace{-0.5em}
    \item \textbf{The least TP $\mu$} is for \textbf{"disgust"} and \textbf{"joy"} emotion with 31 and 48.5 values respectively. "joy" also accounting for the highest TP $\sigma$ with 8.81 value which infers that all the models are agreeing on different datapoints to classify as "joy". Whereas "disgust" has one of the least TP $\sigma$ with 4.0 just following "fear" and "sadness", this suggests that all the models are able to agree on a very small sample space of the class data to be classified as "disgust". 
\end{enumerate}

\section{Conclusion}
In this work, we have explored an application of
BERT and ELECTRA as a means to the task of emotion classification. Various data sampling techniques were used to overcome the large imbalance in data. In the end the best metrics were achieved by using majority voting of the 4 best models as an ensemble. We foresee multiple future directions, including multi-task learning of multiple tasks with a shared backbone, pretraining on the entire GoEmotions dataset, as well as studying and rectifying spurious behaviour of "anger" and "disgust" labels.

\newpage

\bibliography{anthology,custom}
\bibliographystyle{acl_natbib}




\end{document}